\documentclass[conference]{IEEEtran}

\usepackage{graphicx}
\usepackage{caption}
\graphicspath{
    {figures/}
}

\usepackage{amsmath,amssymb,enumerate}

\usepackage{amsthm}
\usepackage{setspace}
\usepackage{booktabs}
\usepackage[usenames,dvipsnames,svgnames,table]{xcolor}
\usepackage{mathtools}
\usepackage{blindtext}
\usepackage{gensymb}
\usepackage{xparse}
\usepackage{lipsum}
\usepackage{mathrsfs}
\usepackage[mathscr]{euscript}
\usepackage{times}
\usepackage[numbers,sort&compress]{natbib}
\usepackage{multicol}
\usepackage{multirow}
\definecolor{darkgreen}{rgb}{0,0.6,0}
\usepackage[bookmarks=true]{hyperref}
\usepackage[caption=false,font=footnotesize]{subfig}
\usepackage{amsfonts}
\usepackage{cleveref}
\usepackage[utf8]{inputenc}
\usepackage[T1]{fontenc}
\usepackage{textcomp}
\usepackage{arydshln}
\usepackage{tabu}
\usepackage{cuted}
\usepackage{flushend}
\usepackage{balance}

\usepackage{thmtools,thm-restate}

\newtheorem{problem}{Problem}

\newtheorem{remark}{Remark}

\newtheorem{definition}{Definition}

\newtheorem{assumption}{Assumption}

\renewcommand{\thefigure}{\arabic{figure}}

\definecolor{note}{rgb}{0.1,0.1,1}
\definecolor{rephase}{rgb}{0.15,0.7,0.15}
\definecolor{bag}{rgb}{0.6,0.6,0.2}

\makeatletter
\renewcommand*\env@matrix[1][c]{\hskip -\arraycolsep
  \let\@ifnextchar\new@ifnextchar
  \array{*\c@MaxMatrixCols #1}}
\makeatother

\newcommand{\transpose}{\mathsf{T}}

\makeatletter
\newcommand{\mathleft}{\@fleqntrue\@mathmargin0pt}
\newcommand{\mathcenter}{\@fleqnfalse}
\makeatother

\usepackage{bm}
\usepackage{cleveref}

\usepackage{amsfonts}
\usepackage{amssymb}
\usepackage{amsbsy}
\usepackage{amsmath}
\usepackage{graphicx}
\usepackage{subcaption}
\usepackage{algorithm}
\usepackage{tablefootnote}
\usepackage{algorithm}
\usepackage{multirow}
\usepackage{algpseudocode}
\usepackage{soul}

\IEEEoverridecommandlockouts

\begin{document}



\title{Riemannian Direct Trajectory Optimization of Rigid~Bodies on Matrix Lie Groups}
\author{Sangli Teng\textsuperscript{1†}, Tzu-Yuan Lin\textsuperscript{1}, William A Clark\textsuperscript{2}, Ram Vasudevan\textsuperscript{1}, and Maani Ghaffari\textsuperscript{1} \\
\textsuperscript{1}University of Michigan, \textsuperscript{2}Ohio University, {†}Corresponding Author: {sanglit@umich.edu} \\ 
}


\makeatletter
\let\@oldmaketitle\@maketitle
    \renewcommand{\@maketitle}{\@oldmaketitle
    \centering
    \includegraphics[width=1.8\columnwidth]{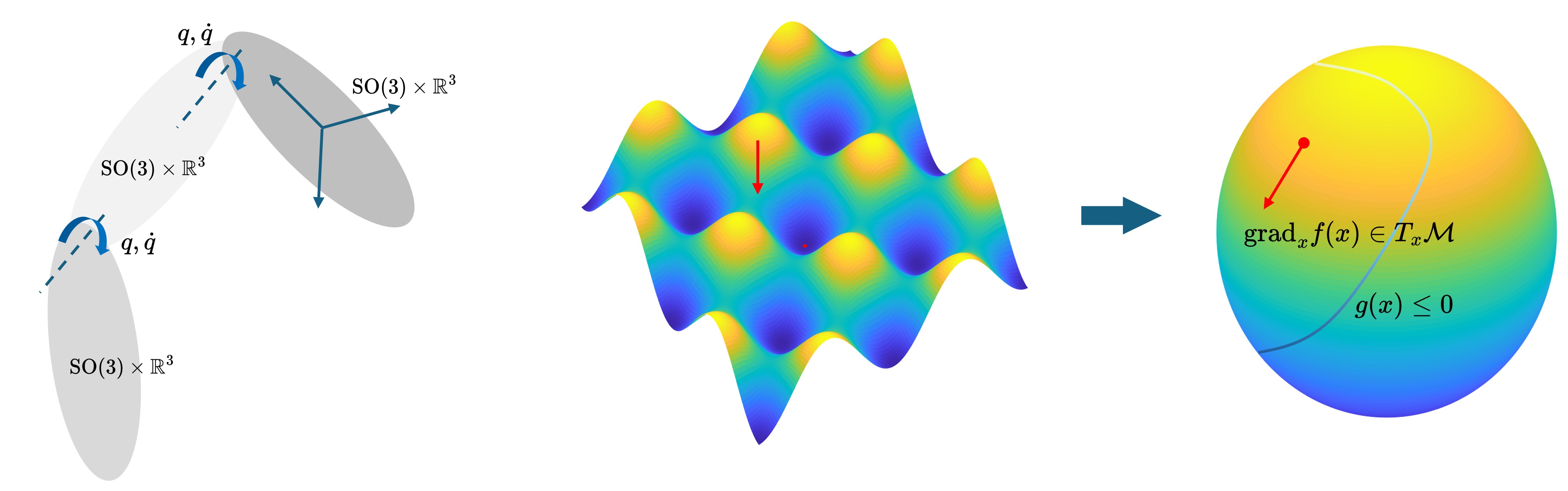}
    \captionof{figure}{Constrained Riemannian direct trajectory optimization of rigid bodies formulated on matrix Lie groups. The rigid body dynamics can be formulated in generalized coordinates ($q \in \mathbb{R}^n$) or on maximal coordinates (product space of $\mathrm{SO}(3)\times \mathbb{R}^3 $). We compare the landscape of a quadratic cost $\|Rz - z\|^2$ defined on $\mathcal{M}=\mathrm{SO}(3)$ projected on $\mathbb{S}^2\cong\mathrm{SO}(3)/\mathrm{SO}(2)$ with different coordinates. With generalized coordinates represented by Euler angles, the landscape is highly complicated with saddle points. With the matrix Lie group coordinates, the decision variable is in a symmetric homogeneous space. In this work, we, for the first time, introduce constrained Riemannian optimization to solve motion planning of rigid bodies modeled on matrix Lie groups.}
    \label{fig:cover}
    \setcounter{figure}{1}
  }
\makeatother

\maketitle
\thispagestyle{plain}
\pagestyle{plain}

\begin{abstract}
    

Designing dynamically feasible trajectories for rigid bodies is a fundamental problem in robotics.
Although direct trajectory optimization is widely applied to solve this problem, inappropriate parameterizations of rigid body dynamics often result in slow convergence and violations of the intrinsic topological structure of the rotation group.
This paper introduces a Riemannian optimization framework for direct trajectory optimization of rigid bodies. 
We first use the Lie Group Variational Integrator to formulate the discrete rigid body dynamics on matrix Lie groups.
We then derive the closed-form first- and second-order Riemannian derivatives of the dynamics. 
Finally, this work applies a line-search Riemannian Interior Point Method (RIPM) to perform trajectory optimization with general nonlinear constraints.
As the optimization is performed on matrix Lie groups, it is correct-by-construction to respect the topological structure of the rotation group and be free of singularities.
The paper demonstrates that both the derivative evaluations and Newton steps required to solve the RIPM exhibit linear complexity with respect to the planning horizon and system degrees of freedom.
Simulation results illustrate that the proposed method is faster than conventional methods by an order of magnitude in challenging robotics tasks.

\end{abstract}




\section{Introduction}


Direct trajectory optimization \cite{hargraves1987direct, betts1998survey, posa2014direct, hereid2017frost} has been extensively applied for motion planning of rigid bodies. Most existing approaches model robot dynamics in generalized coordinates, which evolve on highly complex Riemannian manifolds \cite{bullo1999tracking}. This complexity often leads to slow convergence and computationally expensive derivative evaluations during optimization. On the other hand, as the real projective space $\mathbb{RP}^3$ cannot be embedded differentiably into $\mathbb{R}^3$ \cite{Lee2003}, there does not exist a globally smooth mapping from $\mathbb{R}^3$ to $\mathrm{SO}(3)$ (rotation matrices) or $\mathrm{SU}(2)$ (unit quaternions). Thus, any effort trying to parameterize 3D rotational motions by $\mathbb{R}^3$, such as Euler angles, three Degree-Of-Freedom (DOF) parameterization of quaternion \cite{brudigam2021integrator} or Rodriguez formula \cite{kalabic2017mpc}, breaks the topological structures and introduces singularities.  



The Euler-Poincaré equations provide an alternative formulation \cite{marsden1998introduction, bloch2003nonholonomic}, which describe rigid body dynamics on the matrix Lie group, i.e., a smooth symmetric homogeneous space with superior computational advantages in control~\cite{teng2022input, teng2024convex, 10301632, teng2021toward, liu2025discrete, teng2024generalized, Teng-RSS-23, ghaffari2022progress, teng2022lie, teng2022error} and state estimation~\cite{teng2021legged, yu2023fully, he2024legged, teng2024gmkf}. Despite its potential, a unified framework that fully exploits the Lie group structure for trajectory optimization remains absent. In contrast, \emph{unconstrained} Riemannian optimization \cite{boumal2023introduction} has been widely adopted in robot perception, including Simultaneous Localization and Mapping (SLAM) \cite{forster2016manifold}, sensor registration \cite{clark2021nonparametric}, and globally optimal pose graph optimization \cite{rosen2019se, han2025building}. These perception problems are typically formulated as Maximum A Posteriori estimations that do not involve equality or inequality constraints. However, in motion planning, incorporating constraints such as rigid body dynamics (equality constraints) and collision avoidance (inequality constraints) is necessary to ensure dynamical feasibility.





%

To mitigate this gap, this work bridges the discrete rigid body dynamics on matrix Lie groups and \emph{constrained} Riemannian optimizations to perform \emph{fast} and \emph{topologically compatible} trajectory optimization. The main contribution is illustrated in \Cref{fig:cover} and can be summarized as follows: 
\begin{enumerate}
    \item Derivation of the discrete multi-rigid body dynamics and a unified formulation for direct trajectory optimization of rigid body systems on matrix Lie groups. The proposed formulation is correct-by-construction to preserve the topological structure and conserve the energy.
    \item Derivation of the exact closed-form Riemannian second-order expansion of the dynamics leveraging Lie group symmetry. We show the complexity to evaluate the first- and second-order derivatives is linear w.r.t. the number of rigid bodies.
    \item Development of a line-search Riemannian Interior Point Method to conduct trajectory optimization with general nonlinear constraints. 
    \item Verifications of the proposed method on challenging robotics tasks, including motion planning of drones and manipulators with full dynamics.
    \item Open-source implementation is available at \url{https://github.com/SangliTeng/RiemannianTrajectoryOptimization}.
\end{enumerate}


The remainder of the paper is organized as follows. The related work is summarized in \Cref{sec:related-work}. The math preliminary is provided in \Cref{sec:prelim}. Then we formulate the direct trajectory optimization in \Cref{sec:formulation}. The rigid body dynamics and its differentiation are presented in \Cref{sec:diff-srb-dynamics} and \ref{sec:srb-dynamics}, respectively. A constrained Riemannian optimization is implemented in \Cref{sec:RIPM} and evaluated in \Cref{sec:num-exp}. Finally, the limitations of the proposed method is discussed in \Cref{sec:lim} and the conclusions are summarized in \Cref{sec:conclusion}. 

\section{Related Work}
\label{sec:related-work}
In this section, we review the trajectory optimization of rigid body systems and Riemannian optimizations. 

\subsection{Rigid Body Dynamics}
The majority of robotics applications model the dynamics in generalized coordinates. By expressing the kinetic energy in terms of joint angles, the robot dynamics evolves on the Riemannian manifold with the metric defined by inertia~\cite{bullo2019geometric}. Based on this model, the tracking controller \cite{bullo1999tracking} and its variants in task space \cite{ratliff2018riemannian, sentis2005synthesis, khatib1987unified} can be applied for feedback control of high DOF robots. For optimization-based control, it is more challenging to evaluate the derivatives. The second-order derivative in generalized coordinates has cubic complexity w.r.t. the depth for a kinematic chain, even using the recent state-of-the-art method \cite{10449483}. Highly complicated tensor computations are also expected in this line of research \cite{featherstone2014rigid, featherstone1983calculation, park1995lie}.

Other than modeling the dynamics in generalized coordinates, the maximal coordinates formulation represents each rigid body explicitly and enforces the constraints \cite{leyendecker2008variational, brudigam2021integrator}. The graph structure of the kinematic chain \cite{brudigam2021integrator} preserves the sparsity pattern, thus enabling the computation of gradients to have \emph{linear} complexity w.r.t. the DOF. As each rigid body is explicitly modeled, the manifold structure of the rigid body can be preserved, such as the rigid body simulator \cite{brudigam2021integrator, howelllecleach2022} that applies the variational integrator \cite{marsden2001discrete} in quaternion. However, as \cite{brudigam2021integrator, howelllecleach2022} considers a three-DOF parameterization of quaternion velocities, the representation has singularities. On the other hand, the Lie Group Variational Integrator (LGVI) \cite{lee2005lie} in maximal coordinates \cite{teng2024convex} derives the discrete dynamics of matrix Lie groups that naturally admit a smooth representation that is suitable for Riemannian optimization. 

\subsection{Trajectory Optimization}
Trajectory optimization aims to synthesize robot motions subject to dynamics, kinematics, input, and environment constraints. The direct trajectory optimization derives the dynamics in discrete time and then conducts optimizations \citep{zucker2013chomp, schulman2014motion, posa2014direct, hereid2017frost, manchester2019contact, Dong-RSS-23, li2023autonomous}. By the state-of-the-art numerical optimization techniques \cite{wachter2006implementation, gill2005snopt}, the direct methods can handle large-scale problems with complicated constraints. 


For \emph{rigid body} systems, the single rigid body dynamics has also been applied for tracking control or planning of legged robots \cite{kim2019highly, ding2021representation, agrawal2022vision, teng2022error, grandia2023perceptive}.  The simplified models with dynamics of angular rates neglected are applied in trajectory generation of quadrotors \cite{mueller2013computationally}. However, none of the above methods can synthesize full dynamic trajectories of the robots.  
For full rigid body dynamics, \cite{hereid2017frost, posa2014direct} incorporated the dynamics in generalized coordinates to synthesize the trajectories. The Differential Dynamic Programming (DDP) is applied in \cite{boutselis2020discrete} to synthesize an optimal trajectory on the matrix Lie group. The variation integrator-based dynamics has been applied to the Optimal Control Problem (OCP) of satellite~\cite{junge2005discrete}, while the Lie group version is not fully explored to consider the 3D rotations. The OCP with discrete dynamics on the Lie group is formulated in \cite{kobilarov2011discrete} and can be solved via iterative root finding to meet the first-order optimality condition. The on-manifold Model Predictive Control (MPC) \cite{kalabic2017mpc} synthesized the trajectory with full dynamics on $\mathrm{SO}(3)$ and applied the numerical optimization by lifting the variable to the Lie algebra \cite{lee2005lie}. As \cite{boutselis2020discrete, kobilarov2011discrete, kalabic2017mpc} are designed on Lie groups, they do not face the singularity of gimbal locks when using Euler angles. However, the Rodrigueze formula \cite{kalabic2017mpc}, a three-DOF parameterization of rotational map, inevitably introduces singularities. Though such singularities can sometimes be avoided by manually rounding the velocity vector \cite{wang2025unlocking}, an optimization framework that intrinsically handles the topological structures of the rotation group remains absent. 






\subsection{Riemannian Optimization}
Riemannian optimization has been extensively applied to problems involving variables defined on smooth manifolds \cite{boumal2023introduction}, such as matrix completion \cite{vandereycken2013low} and Semi-Definite Programming (SDP) \cite{wang2023solving, burer2003nonlinear}. The Riemannian derivatives and the retraction map can seamlessly extend traditional first- and second-order numerical optimizations to the manifold setting. Since Riemannian optimization inherently finds search directions on the manifold, it involves far fewer constraints than formulations in the ambient space. For example, the on-manifold Gauss-Newton method has been successfully applied in SLAM \cite{forster2016manifold} on $\mathrm{SE}(3)$. Similarly, optimization on the Stiefel manifold has been proposed to solve the relaxed pose graph optimization to obtain globally optimal solutions \cite{rosen2019se}. Furthermore, Riemannian optimization has demonstrated superior convergence rates in continuous sensor registrations \cite{clark2021nonparametric}.

Beyond perception tasks, OCPs for rigid bodies have been modeled on matrix Lie groups~\cite{teng2024convex, Teng-RSS-23, teng2022lie, teng2022error, 10301632} and could potentially benefit from Riemannian optimization techniques. However, there is a lack of constrained Riemannian solvers tailored for motion planning of rigid bodies on matrix Lie groups. To address this gap, the application of \emph{constrained} Riemannian optimization \cite{schiela2020sqp, obara2022sequential, yamakawa2022sequential, liu2020simple, lai2024riemannian} holds significant promise to consider general nonlinear constraints on matrix Lie groups.




\section{Preliminaries}
\label{sec:prelim}
In this section, we briefly review the basic concepts of Riemannian optimization \cite{boumal2023introduction} and matrix Lie groups. 
\subsection{Riemannian Geometry}
Consider a finite-dimensional smooth manifold $\mathcal{M}$, we denote the tangent space at $x\in \mathcal{M}$ as $\operatorname{T}_x\mathcal{M}$. A vector field is a map $V:\mathcal{M}\rightarrow \operatorname{T}\mathcal{M}$ with $V(x) \in \operatorname{T}_x\mathcal{M}$, and $\operatorname{T} \mathcal{M}:=\bigcup_{x \in \mathcal{M}} \operatorname{T}_x \mathcal{M}$ is the tangent bundle. We denote all smooth vector field on $\mathcal{M}$ as $\mathfrak{X}(\mathcal{M})$. $\mathcal{M}$ is a Riemannian manifold if an Riemannian metric $\langle \cdot, \cdot \rangle_x:\operatorname{T}_x\mathcal{M} \times \operatorname{T}_x\mathcal{M} \rightarrow \mathbb{R}$ for each tangent space $x \in \mathcal{M}$, such that the map 
\begin{equation}
    x \mapsto \langle V(x), U(x) \rangle_x 
\end{equation}
is a smooth function from $\mathcal{M}$ to $\mathbb{R}$. For a function $f:\mathcal{M}_1 \rightarrow \mathcal{M}_2$, the differential is denoted as $\operatorname{D}f(x):\operatorname{T}_x\mathcal{M}_1 \rightarrow \operatorname{T}_x\mathcal{M}_2$. Then we define the gradient and the connection: 

\begin{definition}[Riemannian Gradient]
    Let $f: \mathcal{M} \rightarrow \mathbb{R}$ be a smooth function. The Riemannian gradient of $f$ is the vector field $\operatorname{grad} f$ on $\mathcal{M}$ uniquely defined by the following identities:
    \begin{equation}
        \forall(x, v) \in \mathrm{T}\mathcal{M}, \quad \mathrm{D} f(x)[v]=\langle v, \operatorname{grad} f(x)\rangle_x.
    \end{equation}
\end{definition}

\begin{definition}[Connection]
    An affine connection on a manifold $\mathcal{M}$ is a $\mathbb{R}$-bilinear map $\nabla:\mathfrak{X}(\mathcal{M})\times\mathfrak{X}(\mathcal{M})\to\mathfrak{X}(\mathcal{M})$ such that
    \begin{enumerate}
        \item $\nabla_{fX}Y = f\nabla_XY$, and
        \item $\nabla_X(fY) = \mathcal{L}_Xf\cdot Y + f\nabla_XY$,
    \end{enumerate}
    where $f$ is a smooth function, and $X$ and $Y$ are vector fields.
\end{definition}
Given a Riemannian metric, there exists a unique affine connection, called the Levi-Civita connection, which preserves the metric and is torsion-free; for more details, see e.g. \S6 in \cite{tu2017geometry}. In coordinates, a connection is determined by its Christoffel symbols $\Gamma_{ij}^k$. Let $x = (x^1,\ldots, x^n)$ be local coordinates, then 
\begin{equation}
    \nabla_{\partial_i}\partial_j = \sum_{k} \Gamma_{ij}^k \, \partial_k, \quad \partial_i := \frac{\partial}{\partial x^i}.
\end{equation}
\begin{definition}[Covariant Derivative]
    Let $\mathcal{M}$ be a manifold with connection $\nabla$, and let $c:[a,b]\to \mathcal{M}$ be a smooth curve. Let $X$ be a vector field along the curve $c$. The covariant derivative of $X$ along $c$ is
    \begin{equation}
        \frac{DX}{dt} := \nabla_{\dot{c}}\tilde{X},
    \end{equation}
    where $\tilde{X}$ is any extension of $X$ to the whole manifold.
\end{definition}
The notion of a covariant derivative allows for the definition of a geodesic. A curve, $c(t)$, is a geodesic if the covariant derivative of $\dot{c}$ along $c$ vanishes, i.e.
\begin{equation}
    \ddot{c}(t):=\frac{D\dot{c}}{dt} = 0.
\end{equation}
In coordinates, the geodesic equation is a system of second-order differential equations involving the Christoffel symbols:
\begin{equation}
    \ddot{x}^k + \sum_{ij} \, \Gamma^k_{ij}\dot{x}^i\dot{x}^j = 0, \quad k=1,\ldots, n.
\end{equation}
The Hessian on $\mathcal{M}$ is defined as differentiating the gradients w.r.t. the vector field on $\operatorname{T}{\mathcal{M}}$. Given a tangent vector $u \in \mathrm{T}_x\mathcal{M}$, and a vector field $V$, the derivative of $V$ along $u$ at $x$ is denoted as $\nabla_u V$, given the affine connection $\nabla$ determined by the Riemannian metric $\langle \cdot, \cdot\rangle_x$. Now we proceed to define the Riemannian Hessian:

\begin{definition}[Riemannian Hessian]
Let $\mathcal{M}$ be a Riemannian manifold with its Riemannian connection $\nabla$. The Riemannian Hessian of $f$ at $x \in \mathcal{M}$ is the linear map $\operatorname{Hess} f(x): \mathrm{T}_x \mathcal{M} \rightarrow \mathrm{~T}_x \mathcal{M}$ defined as follows: 
\begin{equation}
    \operatorname{Hess} f(x)[u]=\nabla_u \operatorname{grad} f, \forall u \in \operatorname{T}_x\mathcal{M}.
\end{equation}
\end{definition}
Given the Hessian and gradients at $x \in \mathcal{M}$, one could obtain the search direction $v \in \operatorname{T}_x \mathcal{M} $ in the tangent space for numerical optimization as in Euclidean space. However, moving along the tangent vector $v$ will generally leave the manifold. Thus, we need the retraction map:
\begin{definition}[Retraction]
A retraction on a manifold $\mathcal{M}$ is a smooth map
\begin{equation}
    \operatorname{R}: \mathrm{T}\mathcal{M} \rightarrow \mathcal{M}:(x, v) \mapsto \operatorname{R}_x(v),
\end{equation}
such that each curve $c(t)=\mathrm{R}_x(t v)$ satisfies $c(0)=x$ and $\dot{c}(0)=v$.
\end{definition}
Given the retraction map, we have the second-order Taylor expansion on curves:
\begin{definition}[Second-order Retraction]
\label{def:2nd-retraction}
    Consider $c(t)$ as the retraction curve:
\begin{equation}
    c(t) = \operatorname{R}_x(tv),
\end{equation}
for $ x \in \mathcal{M}$ and $v \in \operatorname{T}_x\mathcal{M}$. We have the second-order retraction:

\begin{equation}
    \begin{aligned}
f\left(\mathrm{R}_x(t v)\right)=& f(x)+t\langle\operatorname{grad} f(x), v\rangle_x 
 +\frac{t^2}{2}\langle \operatorname{Hess} f(x)[v], v\rangle_x \\
& +\frac{t^2}{2}\left\langle\operatorname{grad} f(x), \ddot{c}(0)\right\rangle_x+\mathcal{O}\left(t^3\right) .
\end{aligned}
\end{equation}
\end{definition}

\begin{remark}
\label{rmk:d2c=0}
    In the case that the retraction $\operatorname{R}_x(\cdot) $ is the Riemannian exponential map, the acceleration of $c(t)$, i.e, $\ddot{c}(0) = 0$.  
\end{remark}
\subsection{Matrix Lie Group}
Let $\mathcal{G}$ be an $n$-dimensional matrix Lie group and $\mathfrak{g}$ the associated Lie algebra, i.e, the tangent space of $\mathcal{G}$ at the identity. For convenience, we define the following isomorphism 
\begin{equation}
    (\cdot)^\wedge:\mathbb{R}^n \rightarrow \mathfrak{g}, \quad (\cdot)^\vee:\mathfrak{g} \rightarrow \mathbb{R}^n. 
\end{equation}
that maps between the vector space $\mathbb{R}^n$ and $\mathfrak{g}$. Then, $\forall \phi \in \mathbb{R}^{n}$, we can define the Lie exponential map as
\begin{equation}
    \exp(\cdot):\mathbb{R}^{n} \rightarrow \mathcal{G},\ \ \exp(\phi)=\operatorname{exp_m}({\phi}^\wedge),
\end{equation}
where $\operatorname{exp_m}(\cdot)$ is the exponential of square matrices. We also define the Lie logarithmic map as the inverse of the Lie exponential map:
\begin{equation}
    \log(\cdot): \mathcal{G} \rightarrow \mathbb{R}^{n}. \ \ 
\end{equation}
For every $X \in \mathcal{G}$, the adjoint action, $\mathrm{Ad}_{X}: \mathfrak{g}\rightarrow \mathfrak{g}$, is a Lie algebra isomorphism that enables change of frames:
\begin{equation}
    \mathrm{Ad}_{X}({\phi}^\wedge)= X{{\phi}^\wedge}X^{-1}.
\end{equation}
Its derivative at the identity gives rise to the adjoint map in Lie Algebra as
\begin{equation}
    \mathrm{ad}_{\phi}(\eta) = [{\phi}^\wedge, {\eta}^\wedge],
\end{equation}
where $\phi^\wedge, \eta^\wedge \in \mathfrak{g}$ and $[\cdot, \cdot]$ is the Lie bracket. 

We say a Riemannian metric is left (resp. right) invariant if they are invariant under left (resp. right) group translation:
\begin{equation}
\tag{left-invariant metric}
    \langle X\phi^{\wedge}, X\eta^{\wedge} \rangle_{\operatorname{T}_X\mathcal{G}} = \langle \phi^{\wedge}, \eta^{\wedge} \rangle_{\mathfrak{g}},\ \ 
\end{equation}
\begin{equation}
\tag{right-invariant metric}
    \langle \phi^{\wedge}X, \eta^{\wedge}X \rangle_{\operatorname{T}_X\mathcal{G}} = \langle \phi^{\wedge}, \eta^{\wedge} \rangle_{\mathfrak{g}}.
\end{equation}
A metric is \emph{bi-invariant} if it is both left and right invariant. In general, the Lie exponential and Riemannian exponential are not identical, as bi-invariant metrics may not exist for $\mathcal{G}$.
In particular, for a bi-invariant metric to exist, $\mathrm{Ad}_g$ must be an isometry on $\mathfrak{g}$ \cite{milnor1976}; this is always possible when the group is either compact or Abelian.
In our work, we assume $\mathcal{G}$ is equipped with a bi-invariant metric, which is not restrictive for robotics applications as rigid body motion in $\mathbb{R}^n$ space can be modeled on $\mathrm{SO}(n)\times \mathbb{R}^n$, a product space of a compact and an Abelian group: 
\begin{assumption}
\label{assumption:Lie=Rieman}
    We assume the group we study, namely $\mathcal{G}$, admits a bi-invariant metric, and the Riemannian exponential is identical to the Lie exponential in the following derivation. 
\end{assumption}


\section{Problem Formulations}
\label{sec:formulation}
Now we formally define the direct trajectory optimization of rigid bodies in 3D spaces. For a rigid body system composed of $N_b$ single rigid bodies modeled on $\mathrm{SO}(3)\times \mathbb{R}^3$, we define the configuration space as the product space:
\begin{equation}
    \mathcal{M}_{\mathrm{RB}}:= \underbrace{ (\mathrm{SO}(3)\times \mathbb{R}^3) \times \cdots \times (\mathrm{SO}(3)\times \mathbb{R}^3)  }_{N_b}.
\end{equation}
We define the time step $\Delta t \in \mathbb{R}$ and the time sequence \mbox{$\{t_k = k\Delta t \mid  k = 0, \dots, N\} \subset \mathbb{R}$}. We assume that the set of all feasible control inputs is $u \in \mathcal{U} \subseteq \mathbb{R}^m$, and we have the implicit rigid body dynamics:
\begin{equation}
    f_d(x_{k+1}, x_k, u_{k}) = 0, x \in \mathcal{M}_{RB}, u \in \mathcal{U},
\end{equation}
that relates the discrete configuration states at time $t_k$ and $t_{k+1}$. Then we have the optimization:
\begin{problem}[Direct Trajectory Optimization of Rigid Bodies]
\label{prob:traj-opt}
    Consider the configuration space $\mathcal{M}_{\mathrm{RB}}$ of a $N_b$ rigid body system and the set of feasible control input $\mathcal{U}$, we synthesize the optimal trajectory of $x \in \mathcal{M}_{\mathrm{RB}}$ via the optimization:
    \begin{equation}
        \begin{aligned}
            \min_{ \{x_k\}_{k=0}^{N}, \{u_k\}_{k=0}^{N-1} } \quad &P(x_N) + \sum_{k=0}^{N-1} L(x_k, u_{k}) \\
            \mathrm{s.t.} \quad & f_d(x_{k+1}, x_k, u_{k}) = 0, \\
                                & g(x_{k+1}) \le 0, \\
                                & u_{k} \in \mathcal{U}, x_{k+1} \in \mathcal{M}_{\mathrm{RB}} \\
                                & k = 0, 1, 2, \cdots, N-1, \\
                                & x_0 = x_{\mathrm{init}}.\\
        \end{aligned}
    \end{equation}
    with $L(\cdot, \cdot)$ the stage cost, $P(\cdot, \cdot)$ the terminal cost, $g(\cdot)$ the inequality constraints, and $x_{\mathrm{init}}$ the initial condition. 
\end{problem}
As the state $x \in \mathcal{M}_{\mathrm{RB}}$ admits manifold structure on matrix Lie groups, in the following sections, we apply Riemannian optimization to solve \Cref{prob:traj-opt}. The key to this goal is to formulate the discrete dynamics on matrix Lie groups and derive their Riemannian derivative. 
\section{Discrete Rigid Body Dynamics}
\label{sec:srb-dynamics}
In this section, we derive the rigid body dynamics based on the variational integrators on $\mathcal{M}_{\mathrm{RB}}$. 

\subsection{Variation-based Discretization}

Consider a mechanical system with the configuration space $\mathcal{M}$. We denote the configuration state as $x \in \mathcal{M}$ and the generalized velocity as $\dot{x} \in \operatorname{T}_{x}\mathcal{M}$. Then we have the Lagrangian given the kinetic and potential energy $T(\dot{x}), V(x)$:
\begin{equation}
\label{eq:ct_lag}
    L(x, \dot{x}):=T(\dot{x}) - V(x). 
\end{equation}
The key idea of a variational integrator is to discretize the Lagrangian \eqref{eq:ct_lag} to obtain the discrete-time EoM \cite{marsden2001discrete}. The discretization scheme ensures that the Lagrangian is conserved in discrete time, thus having superior energy conservation properties over long durations. 
The discrete Lagrangian $L_d: \mathcal{M}\times \mathcal{M} \rightarrow \mathbb{R}$ could be considered as the approximation of the action integral via:
\begin{equation}
\label{eq:lag_ct2dt}
    L_d(x_k, x_{k+1}) \approx \int ^{t_{k+1}}_{t_k}L(x, \dot{x})dt .
\end{equation}
Then the discrete variant of the action integral becomes:
\begin{equation}
    S_d = \sum_{k=0}^{N-1}{L}_d(q_k, q_{k+1}) .
\end{equation}
Finally, we take variation in $\operatorname{T}\mathcal{M}$ and group the term corresponding to $\delta x_{k} \in \operatorname{T}_{x_k}\mathcal{M}$ as the discrete version of integration by parts \cite{marsden2001discrete}: 
\begin{equation}
\begin{aligned}
  \delta S_d &= \langle D_1 {L}_d(x_0, x_1), \delta x_0 \rangle +  \langle D_2 {L}_d(x_{N-1}, x_N),  \delta x_N \rangle\\ 
    &+ \sum_{k=1}^{N-1}
    \langle D_2 {L}_d(x_{k-1}, x_k) + D_1 {L}_d(x_k, x_{k+1}), \delta x_k \rangle. 
\end{aligned}
\end{equation}
where $D_i$ denotes the derivative with respect to the $i$-{th} argument and $\langle \cdot, \cdot\rangle$ the canonical inner product.  
By the least action principle, the stationary point can be determined by letting the derivative of $\delta x_k$ be zero: 
\begin{equation}
\label{eq:discrete_eom}
D_1L_d(x_k, x_{k+1}) + D_2L_d(x_{k-1}, x_{k}) = 0. 
\end{equation}
For the holonomic constraints specified by the manifold:
\begin{equation}
\label{eq:holonomic-cons}
    h(x) = 0 \in \mathbb{R}^m,
\end{equation}
another action integral can be considered:
\begin{equation}
\begin{aligned}
    \int_{t_k}^{t_{k+1}} \langle \lambda(t), h(x)\rangle dt &\approx \frac{\Delta t}{2} \langle\lambda(t_k), J(x_k)\delta x_k\rangle  \\
    & + \frac{\Delta t}{2} \langle \lambda(t_{k+1}), J(x_{k+1})\delta x_{k+1}\rangle. 
\end{aligned}
\end{equation}
with \begin{equation}
\label{eq:cons-Jac}
    J(x_k) := \frac{\partial h(x_k)}{\partial x_k}
\end{equation} the Jacobian of the constraints, $\delta h(x_k) = J(x_k)\delta x_k$, and $\lambda$ the constrained force.
By adding these terms to the unconstrained Lagrangian and taking variation w.r.t $x_k$, we have the constrained dynamics:
\begin{align}
\label{eq:dynamics-full}
    D_1L_d(x_k, x_{k+1}) + D_2L_d(x_{k-1}, x_{k}) &= J(x_k)^{\transpose}\lambda_k \Delta t, \\
    h(x_{k+1}) &= 0.
\end{align}
The constraints $h(x_{k+1}) = 0$ are needed to ensure the states in discrete time do not leave the manifold, which is essential in the multi-body case. 
To consider the external force $u$ we consider the action integral of control input $u(t)$:
\begin{equation}
    \int_{t_k}^{t_{k+1}} \langle u(t), \delta x \rangle dt \approx \frac{\Delta t}{2} \langle u(t_k), \delta x_k\rangle + \frac{\Delta t}{2} \langle u(t_{k+1}), \delta x_{k+1} \rangle. 
\end{equation}

\subsection{Dynamics on $\mathrm{SO}(3)\times \mathbb{R}^3$}
Now, we derive the EoM on $\mathrm{SO}(3)\times \mathbb{R}^3$. Consider the discrete equation of motion:
\begin{equation}
\label{eq:SO3R3-kin}
\begin{aligned}
    R_{k+1} &= R_kF_k \in \mathrm{SO}(3), \\
    p_{k+1} &= p_k + v_k \Delta t,
\end{aligned}
\end{equation}
with $R_k$ the orientation, $F_k$ the discrete pose change, $p_k$ the position, and $v_k$ the linear velocity. The mid-point approximation can be applied:
\begin{equation}
\label{eq:Fk_midpoint}
\begin{aligned}
    F_k := R_k^{-1}R_{k+1} \approx I + \Delta t\omega_k^{\times}, \ \ \omega_k^{\times} \approx \frac{F_k - I}{\Delta t},
\end{aligned}
\end{equation}
\begin{equation}
\label{eq:pk_midpoint}
\begin{aligned}
    \dot{p}_k = v_k \approx \frac{p_{k+1} - p_k}{\Delta t}.
\end{aligned}
\end{equation}
The kinetic and potential energy can be approximated by:
\begin{align}
    \nonumber T_d&:=\frac{1}{2\Delta t}\operatorname{tr}((F_k - I)I^b(F_k - I)^{\transpose}) + \frac{1}{2\Delta t}m\|p_{k+1} - p_k\|^2, \\
    V_d&:= m\left(\frac{p_{k+1}+p_{k}}{2}\right)^{\transpose} g \Delta t,
\end{align} 
where and $I^b$ the nonstandard moment of inertia \citep{marsden1999discrete} that relate the standard moment of inertia $I_b$ by $I_b = \operatorname{tr}(I^b)I_3-I^b$. Via the taking variation on $\mathrm{SO}(3) \times \mathrm{R}^3$, we have the dynamics:
\begin{align}
\label{eq:lgvi_dynamics}
    &F_{k+1}I^b - I^bF_{k+1}^{\transpose}=I^bF_{k} - F_k^{\transpose}I^b, \\ 
    &mv_{k+1} = mv_{k} + mg\Delta t. 
\end{align}
With the constraints formulated on $\mathrm{SO}(3)\times \mathbb{R}^3$, we can obtain the dynamics for multi-body systems. Compared to an explicit integration scheme, the LGVI naturally obeys the manifold constraints and conserves the energy \cite{marsden2001discrete, lee2005lie}. As the LGVI is completely in matrix form, there is no need to move back and forth between the Lie group and its Lie algebra for integration. The reader can refer to \cite{teng2024convex} for a detailed comparison of integrators on Lie groups for motion planning,  

\section{Differentiate the Rigid Body Dynamics}
\label{sec:diff-srb-dynamics} 
In this section, we derive the first- and second-order Riemannian derivatives of kinematics and dynamics of rigid body motion. Under \Cref{assumption:Lie=Rieman} and remembering \Cref{rmk:d2c=0}, the BCH formula is sufficient to obtain the Riemannian gradients and Hessians via the retraction in \Cref{def:2nd-retraction}.

\begin{table*}
\centering
\caption{Summary of equations of motions for discrete rigid body dynamics and conventional constraints.}
{\renewcommand{\arraystretch}{2}
\begin{tabular}{ccc}
\hline
 & Constraints & Second-order Expansions \\ \hline
$\begin{array}{c}
    \text{Rotational} \\
    \text{Kinematics}
\end{array}$
&  $\log{Y}:=\log{( R_{k+1}^{-1}R_kF_k )} = 0 \in \mathfrak{so}(3)$   & $\begin{aligned}
&-Y^{-1}\xi^{R}_{k+1} + F_k^{-1}\xi^R_k + \xi_k^F + \\
&0.5[-Y^{-1}\xi^R_{k+1}, F_k^{-1}\xi^R_{k} ] + 0.5[-Y^{-1}\xi^R_{k+1},\xi^F_{k} ] \\ 
&+0.5[F_k^{-1}\xi_k^R,\xi^F_{k} ] \\
\end{aligned}$ \\ \hline
$\begin{array}{c}
    \text{Translational} \\
    \text{Kinematics}
\end{array}$  & $p_{k+1} - p_k - v_k\Delta t = 0  \in \mathbb{R}^3$ & $\xi^p_{k+1} - \xi_k^p - \xi_k^p\Delta t$ \\ \hline 
$\begin{array}{c}
    \text{Rotational} \\
    \text{Dynamics}
\end{array}$ &  $ \begin{aligned}
    &(F_{k+1}I^b - I^bF_{k+1}^{\transpose} - (I^bF_{k} - F_k^{\transpose}I^b))^{\vee} = 0 \in \mathfrak{so}(3)
\end{aligned} $   & $\begin{aligned}
     &I_bF_{k+1}\xi^F_{k+1} - I_bF_k\xi^F_k \\
    +&0.5(F_{k+1}\xi_{k+1}^{F\wedge 2}I^b - I^b\xi_{k+1}^{F\wedge 2}F_{k+1}^{\transpose})^{\vee} \\
    +&0.5(I^bF_{k}\xi_{k}^{F\wedge 2} - \xi_{k}^{F\wedge 2}F_k^{\transpose}I^b)^{\vee} \\
\end{aligned} $ \\ \hline
$\begin{array}{c}
    \text{Translational} \\
    \text{Dynamics}
\end{array}$  & $mv_{k+1} - mv_k - mg\Delta t = 0 \in \mathbb{R}^3$ & $m\xi^v_{k+1} - m\xi^v_k$ \\ \hline 

$\begin{array}{c}
    \text{Pivot} \\
    \text{Constraints}
\end{array}$  & $R_1r_1 + p_1 - (R_2r_2 + p_2) = 0 \in \mathbb{R}^3$ &  $\begin{aligned}
    &R_1(\xi^{R\wedge}_1 + 0.5\xi^{R\wedge 2}_1 )r_1 + \xi_1^{p} - \\
    & \quad R_2(\xi^{R\wedge}_2 + 0.5\xi^{R\wedge 2}_2 )r_2 + \xi_2^{p} \\
\end{aligned}$ \\ \hline 

$\begin{array}{c}
    \text{Axis}\\
    \text{Constraints}
\end{array}$  & $(R_1e_{i})^{\transpose}(R_2e_z) = 0, i = x, y$ &  $\begin{aligned}
    -&(R_1e_{i})^{\transpose}(R_2  e_z^{\wedge})\xi_2^{R\wedge} - (R_2e_{z})^{\transpose}(R_1  e_i^{\wedge})\xi_1^{R\wedge} \\
    & \quad + 0.5(R_1e_{i})^{\transpose}(R_2 \xi_2^{R\wedge} e_z) \\
    \vspace{1pt}
    & \quad + 0.5(R_2e_{z})^{\transpose}(R_1 \xi_1^{R\wedge} e_i) \\
\end{aligned}$ \\  
\hline 
\end{tabular}}
\label{table:cons_2nd_retraction}
\end{table*}

\subsection{Differentiate the Kinematic Chain}
Without loss of generality, we consider the kinematic chain constraints on $\mathcal{G}$ with finite length $n$:
\begin{equation}
\label{eq:cons_kin}
    X_1X_2\cdots X_{n-1}X_n = I \in \mathcal{G}.
\end{equation}
Now we leverage the BCH formula to derive the second-order retraction at the operating point $\bar{X}_1, \bar{X}_2, \cdots, \bar{X}_n$. We denote $\bar{Y}:=\bar{X}_1\bar{X}_2 \cdots \bar{X}_N$. To avoid operating points other than $I\in \mathcal{G}$, we reformulated constraints \eqref{eq:cons_kin} as: 
\begin{equation}
\begin{aligned}
&\bar{Y}^{-1}X_1X_2\cdots X_{n-1}X_n = \bar{Y}^{-1},
\end{aligned}
\end{equation}
and vectorize it via the logarithmic map:
\begin{equation}
\label{eq:vec-kin}
    \log{(\bar{Y}^{-1}X_1X_2\cdots X_{n-1}X_n)} = \log{(\bar{Y}^{-1})}. 
\end{equation}
Consider the tangent vector $ \bar{X}_i \xi_i^{\wedge} \in \operatorname{T}_{\bar{X}_i} \mathcal{G}$, the retraction by the Riemannian exponential under \Cref{assumption:Lie=Rieman} is:
\begin{equation}
   c_i(t) = \bar{X}_i \exp{(t\xi_i)}.
\end{equation}
With $X_{i, j} = X_i X_{i+1} \cdots X_j, i \le j$ and $X_{i+1, i} = I$, the second-order retraction of \eqref{eq:vec-kin} can be derived as:
\begin{equation}
     \begin{aligned}
         &\log{(\bar{Y}^{-1}X_1\exp{(t\xi_1)}\cdots X_{n-1}\exp{(t\xi_{n-1})}X_n\exp{(t\xi_{n})})} =  \\
         & t\sum_{i=1}^n \operatorname{Ad}_{X_{i+1, n}^{-1}} \xi_i + \frac{t^2}{2}\sum_{i < j} [\operatorname{Ad}_{X_{i+1, n}^{-1}} \xi_i, \operatorname{Ad}_{X_{j+1, n}^{-1}} \xi_j] +  \mathcal{O}(t^3)
     \end{aligned}
\end{equation}
The Riemannian gradient and Hessian can be obtained by evaluating the first- and second-order terms of $t$. For clarity, the detailed derivation is deferred to Appendix \ref{appx:kin-derivation}. 

\subsection{Differentiate the Dynamics and Constraints}
Consider the rotational dynamics in \eqref{eq:lgvi_dynamics}, we again use the the retraction curve along the direction $\xi^{F}_i \in \mathfrak{so}(3)$:
\begin{equation}
    c(t) = \bar{F}_i\exp{(t\xi^{F}_i)} \in SO(3). 
\end{equation}
As the \eqref{eq:lgvi_dynamics} is already a skew matrix, the vectorized perturbed dynamics can be obtained by substituting the retraction curve and applying the $(\cdot)^{\vee}$ map:
\begin{equation}
\begin{aligned}
    &(F_{k+1}\exp{(t\xi^{F}_{k+1})} I^b - I^b\exp{(-t\xi^{F}_{k+1})}F_{k+1}^{\transpose})^{\vee}= \\
    & \quad (I^bF_{k}\exp{(t\xi^{F}_{k})} - \exp{(-t\xi^{F}_{k})}F_k^{\transpose}I^b)^{\vee}. 
\end{aligned}
\end{equation}
Via the Taylor expansion of the exponential map:
\begin{equation}
    \exp{(t\xi)}\approx I + \frac{ t\xi^{\wedge} }{1!} + \frac{ t^2\xi^{\wedge2} }{2!} + \mathcal{O}(t^3),
\end{equation}
the second-order expansion can be obtained by substituting the series and keeping the first and second-order terms of $t$.

Then we apply the second-order retraction again to differentiate the holonomic constraints. Here, we consider two common constraints: pivot constraints and axis constraints. Given two rigid bodies connected by a pivot joint, we have:
\begin{equation}
    R_1r_1 + p_1 = R_2r_2 + p_2 \tag{Pivot Constraints},
\end{equation}
where $R_i$ is the orientation of the rigid body in the world frame, $p_i$ is the position of the rigid body, and $r_i$ is the position vector pointing from the center of mass (CoM) of each rigid body to the joint, represented in each rigid body frame. 
The associated second-order retraction can be computed from:
\begin{equation}
    R_1\exp{(t\xi^{R}_1)}r_1 + (p_1 + t\xi^{p}_2) = R_2\exp{(t\xi^{R}_2)}r_2 + (p_2 + t\xi^{p}_2). 
\end{equation}
To constrain the rotation to one axis, e.g, $e_z$, we have:
\begin{equation}
    \begin{aligned}
        (R_1e_{i})^{\transpose}(R_2e_z) = 0, i = x, y,
    \end{aligned} \tag{Axis Constraints}
\end{equation}
where $e_i$ is the unit vector of each axis. Then, the second-order retraction can be computed from:
\begin{equation}
    (R_1\exp{(t\xi^{R}_1)}e_{i})^{\transpose}(R_2\exp{(t\xi^{R}_2)}e_z) = 0, i = x, y,
\end{equation}
Finally, we summarize all the constraints and the corresponding second-order retraction in \Cref{table:cons_2nd_retraction}.

\subsection{Complexity of Differentiation}
\begin{figure*}
    \centering
    \includegraphics[width=.8\linewidth]{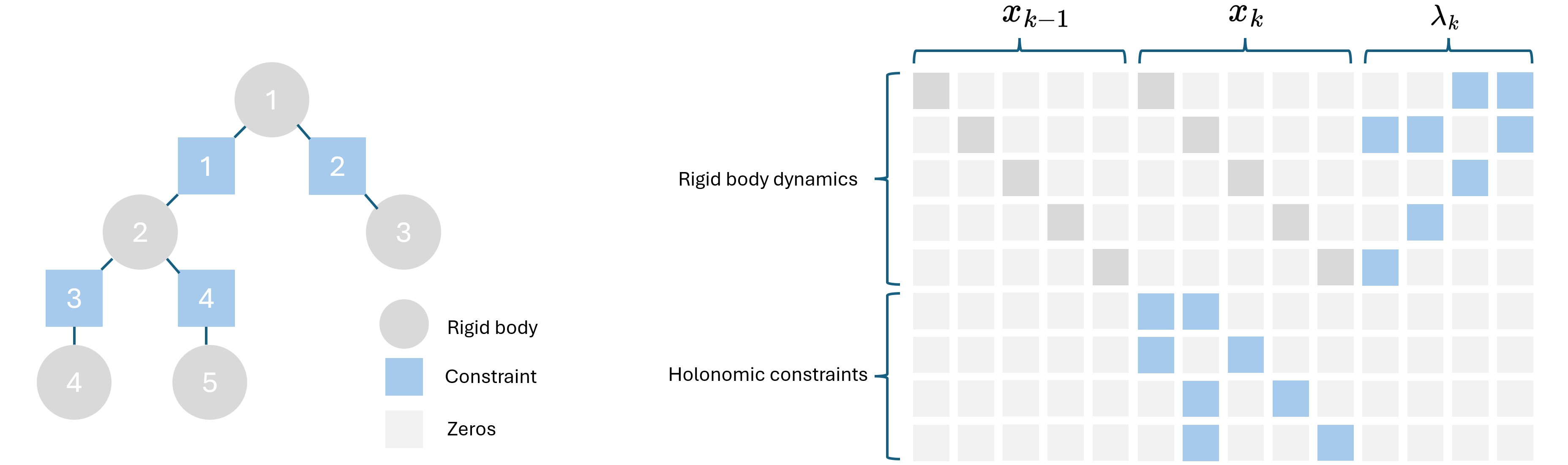}
    \caption{Example of a rigid body system in maximal coordinates and its Jacobian. The blue block indicates the Jacobians with nonzero elements. The lighter gray box indicates zero block, while the darker gray box indicates the Jacobians of the autonomous single rigid body dynamics.}
    \label{fig:max-mrb}
\end{figure*}

As the maximal coordinate formulation admits a graph structure \cite{brudigam2021integrator}, we conclude that the complexities of evaluating the first- and second-order derivatives are linear w.r.t. the number of rigid bodies and kinematic pairs. The derivative for each rigid body can be evaluated by the first four rows of \Cref{table:cons_2nd_retraction}. As the kinematic pair only involves the configuration state of two rigid bodies, the Hessian only involves the variables involved in the two rigid bodies. An example of the Jacobian is illustrated in \Cref{fig:max-mrb}.

\section{Constrained Riemannian Optimization}
\label{sec:RIPM}
In this section, we apply the Riemannian Interior Point (RIPM) to conduct trajectory optimization. Though an RIPM is implemented in \cite{lai2024riemannian}, we provide a customized implementation without dependencies on Manopt \cite{boumal2014manopt}. We mainly refer to IPOPT \cite{wachter2006implementation} for the line-search implementation.



\subsection{Riemannain KKT Condition}
We introduce the optimality condition of constrained optimization defined on smooth manifolds. By the language of Riemannian geometry, the condition coincides with its counterparts in Euclidean space \cite{yang2014optimality}.  
Consider an optimization problem with general nonlinear constraints:
\begin{problem}[Constrained Riemannian Optimization]
\label{prob:cons-nlp}
\begin{equation}
\begin{aligned}
    \min_{x \in \mathcal{M} } \quad &f(x) \\
    &h_i(x) = 0, i = 1, 2, \cdots, l, \\
    &g_j(x) \le 0, j = 1, 2, \cdots, m.
\end{aligned}    
\end{equation}
with the variable $x$ defined on smooth manifold $\mathcal{M}$, $\{h_i(x)\}_{i=1}^{l}$ the equality constraints and $\{g_i(x)\}_{i=1}^m$ the inequality constraints. 
\end{problem}

With the multiplier $y \in \mathbb{R}^{l}$ and $z \in \mathbb{R}^m$, we further have the Lagrangian defined on the product manifold $\mathcal{M}\times \mathbb{R}^l \times \mathbb{R}^m$:
\begin{equation}
    \mathcal{L}(x, y, z) := f(x) + \sum_{i=1}^{l} y_i h_i(x) + \sum_{i=1}^{m} z_i g_i(x).
\end{equation}
The Riemannian gradients can then be obtained:
\begin{equation}
\begin{aligned}
    \operatorname{grad}_x\mathcal{L}(x, y, z) = &\operatorname{grad}_x f(x) + \sum_{i=1}^{l} y_i \operatorname{grad}_x h_i(x)\\ + &\sum_{i=1}^{m} z_i \operatorname{grad}_x g_i(x)
\end{aligned}
\end{equation}
Then we have the following optimality conditions:
\begin{definition}[First-Order Optimality Conditions]
\label{theorem:kkt}
    $\hat{x} \in \mathcal{M} $ is said to satisfy the KKT conditions if exist the multipliers $\hat{y}$ and $\hat{z}$, such that:
    \begin{enumerate}
        \item Stationary condition: $\operatorname{grad}_x\mathcal{L}(\hat{x}, \hat{y}, \hat{z}) = 0$,
        \item Primal feasibility: $h_i(\hat{x}) = 0, g_j(\hat{x})\ge 0, \forall i, j$,
        \item Dual feasibility: $\hat{z}_j \ge 0, \forall j$,
        \item Complementarity condition: $\hat{z}_jg_j(\hat{x}) = 0, \forall j$.
    \end{enumerate}
    
\end{definition}
We then proceed to introduce the RIPM to search for the critical point that satisfies the KKT condition.

\subsection{Riemannian Interior Point Method}
In our application, we apply a line-search RIPM to find the KKT pair $(\hat{x}, \hat{y}, \hat{z}) \in \mathcal{M}\times \mathbb{R}^l \times \mathbb{R}^m$ that satisfies the optimality condition in \Cref{theorem:kkt}. We consider the log-barrier problem with homotopy parameter $\mu$ and slack variables $s$:
\begin{problem}[Log-barrier Problem]
    \begin{equation}
\begin{aligned}
    \min_{x \in \mathcal{M} } \quad \varphi_{\mu} := &f(x) - \mu \sum_{i=1}^{m} \log{s_i} \\
    &h_i(x) = 0, i = 1, 2, \cdots, l, \\
    &g_j(x) + s_j = 0, j = 1, 2, \cdots, m, \\
    & s \ge 0.
\end{aligned}
\end{equation}
\end{problem}
The convergence to \Cref{prob:cons-nlp} ($\varphi_0$) can be guaranteed if $\mu \rightarrow 0$ \cite{wright2006numerical}. The KKT condition of the barrier function can be written as:
\begin{equation}
\begin{aligned}
    \operatorname{grad}_x f(x) + A_\mathrm{E}(x) y  + A_\mathrm{I}(x) z &= 0 \\ 
    h(x) &= 0 \\
    g(x) + s &= 0 \\
    Sz - \mu e &= 0
\end{aligned}
\end{equation}
with the Jacobian evaluated at $x \in \mathcal{M}$ defined as:
\begin{equation}
    A_{\mathrm{E}}(x): \mathbb{R}^{l} \rightarrow \operatorname{T}_x\mathcal{M}:  A_{\mathrm{E}}(x)y := \sum_{i=1}^{l} y_i \operatorname{grad}_x h_i(x),
\end{equation}
\begin{equation}
    A_{\mathrm{I}}(x): \mathbb{R}^{m} \rightarrow \operatorname{T}_x\mathcal{M}: A_{\mathrm{I}}(x)z := \sum_{i=1}^{m} z_i \operatorname{grad}_x g_i(x).
\end{equation}
Then we differentiate the KKT vector field:
\begin{equation}
\label{eq:KKT-log-barrier}
\begin{aligned}
    \left[\begin{array}{cccc}
\operatorname{Hess}_x \mathcal{L}(x, y, z) & A_{\mathrm{E}}^*(x) & A_{\mathrm{I}}^*(x)  & 0 \\
A_{\mathrm{E}}(x) & 0 & 0 & 0 \\
A_{\mathrm{I}}(x) & 0 & 0 & I \\
0 & 0 & S & Z
\end{array}\right]\left[\begin{array}{c}
d_x \\
d_y \\
d_z \\
d_s
\end{array}\right]= \\ 
-\left[\begin{array}{c}
\nabla f(x)+A_{\mathrm{E}}^*y +A_{\mathrm{I}}^*z \\
h(x) \\
g(x)+s \\
S z-\mu e \\
\end{array}\right]
\end{aligned} 
\end{equation}
with $(d_x, d_y, d_z, d_s) \in \operatorname{T}_x\mathcal{M} \times \mathbb{R}^l \times \mathbb{R}^m \times \mathbb{R}^m $ the search direction, $Z = \operatorname{diag}(z)$,  $S = \operatorname{diag}(s)$ and {$A^{*}$ the adjoint of $A$}. The Hessian of the Lagrangian can be obtained by:
\begin{equation}
\begin{aligned}
    \operatorname{Hess}_x\mathcal{L}(x, y, z) = &\operatorname{Hess}_x f(x) + \sum_{i=1}^{l} y_i \operatorname{Hess}_x h_i(x)\\ + &\sum_{i=1}^{m} z_i \operatorname{Hess}_x g_i(x).
\end{aligned}
\end{equation}

After obtaining the search direction via Riemannian Newton's method according to \eqref{eq:KKT-log-barrier}, we implement a backtracking line-search method to decide the step size \cite{wachter2006implementation}. The procedure is summarized in \Cref{alg:line-search-RIPM}. The logic of \Cref{alg:line-search-RIPM} is to accept the Newton step only if the improvement in feasibility or the reduction in cost is sufficient. The condition for adjusting the homotopy parameter for the log-barrier problem is determined by the violation of the optimality condition $E_{\mu}$:
\begin{equation}
E_{\mu} = \max\left\{ \epsilon_{\mathrm{KKT}}, \epsilon_{\mathrm{E}},
\epsilon_{\mathrm{I}}, \right\},
\end{equation}
with the violation of the KKT condition, equality constraints and inequality constraints being
\begin{equation}
\begin{aligned}
\epsilon_{\mathrm{KKT}} &= \frac{\| \operatorname{grad}_xf(x)+A_{\mathrm{E}}y + \mathcal{A}_Iz \|_{\infty}}{s_d}, \\
\epsilon_{\mathrm{E}} &= \| h(x) \|_{\infty}, \ \ \epsilon_{\mathrm{I}} = \frac{\| Sz - \mu e \|_{\infty}}{s_c},
\end{aligned}
\end{equation}
respectively. The $s_d$ and $s_c$ are normalizing parameters, and the termination condition for \Cref{prob:cons-nlp} is $E_{0}$. The parameters are shown in \Cref{table:param-RIPM} in the Appendix. \ref{appx:ripm-table}.

\begin{algorithm}[t]
\label{alg:tro-single}

\caption{Line-Search Riemannian Interior Point Method}\label{alg:line-search-RIPM}
\begin{algorithmic}
\footnotesize
\Require Initial guess $(x_1, y_1, z_1, s_1) \in \mathcal{M} \times \mathbb{R}^l \times \mathbb{R}^m \times \mathbb{R}^m$, homotopy parameter $\mu$, parameters in \Cref{table:param-RIPM}

\For{$k = 1, \ldots, N_{\max}$}
    \State {\texttt{// Decide the search direction}}
    \State $(d_k^x, d_k^{y}, d_k^z, d_k^s) \gets $ solving \eqref{eq:KKT-log-barrier} evaluated at $(x_k, y_k, z_k, s_k)$

    \State {\texttt{// Check the termination condition}}
    \If {$E_0(d_k^x, d_k^{y}, d_k^z, d_k^s) \le \epsilon_{\mathrm{tol}}$}
        \State \textbf{break}
    \EndIf

    \State {\texttt{// Update the homotopy parameter}}     
    \If {$E_\mu(d_k^x, d_k^{y}, d_k^z, d_k^s) \le 10\mu$} 
        \State $\mu \gets \max\{ \frac{\epsilon_{\mathrm{tol}}}{10}, \min \{ \kappa_{\mu}\mu, \mu^{\theta_{\mu}} \} \}$ 
    \EndIf

    \State {\texttt{// Step size by fraction to boundary rule}}
    \State $\tau \gets \max \left\{\tau_{\min}, 1-\mu\right\}$
    \State $\alpha_k^z \gets \max\{ \alpha \in (0, 1] \ | \ z + \alpha d^z_k \ge (1 - \tau_j) z \} $
    \State $\alpha_k^s \gets \max\{ \alpha \in (0, 1] \ | \ s + \alpha d^s_k \ge (1 - \tau_j) s \} $

    \State $d_k^{z} \gets \alpha_k^{z} d_k^{z} $, $d_k^{s} \gets \alpha_k^{s} d_k^{s} $

    \State {\texttt{// Backtracking line-search}}
    \State $ \varphi_{\mu, k}^* \gets \varphi_{\mu,}(x_k) $
    \State $\theta_k^* \gets \|h(x_k)\|_1 + \sum_{i, g_i(x_k) \ge 0} g_i(x_k) $
    
    \For {$j = 1, \ldots, J_{\max}$}
        \State $\Tilde{x} \gets \operatorname{R}_{x_k}(d^x_k)$ \textcolor{black}{\emph{// Apply the retraction map}} 

        \State {\texttt{// Log barrier loss and constraints violation }}
        \State $\Tilde{\varphi_{\mu}} \gets \varphi_{\mu}(\Tilde{x}) $, $\Tilde{\theta} \gets \|h(\Tilde{x})\|$, $c \gets \langle \operatorname{grad}_{x} f(\Tilde{x})  , d_k^x \rangle$


        \If {$\theta^*_k \le \theta_{\min}$ \textbf{and} $c<0$ \textbf{and} $\alpha_{k, j}c^{s_{\varphi}} > \delta \theta^{*s_{\theta}} $}
        \If { $\varphi_{\mu}(\Tilde{x}) \le \varphi_{\mu}(x_k) + \eta_{\varphi}c $ }
        \State \textbf{break} {\texttt{// Armijo condition satisfied}}
        \EndIf
        
        \ElsIf {$\Tilde{\varphi}_{\mu} \le {\varphi}_{\mu, k}^* - \gamma_{\theta} {\theta}_k^* $ \textbf{or} {$\Tilde{\theta} \le (1 - \gamma_{\theta}) {\theta}_k^* $}}
        \State \textbf{break} {\texttt{// Feasibility or cost is improved }}
        \EndIf
        
        \State $d_k^x \gets \beta d_k^x$ {\texttt{// Update step size}}
    \EndFor
    \State {\texttt{// Update the variables}}
    \State $x_{k+1} \gets \Tilde{x}$, $y_{k+1} \gets y_k + d_k^y$, $z_{k+1} \gets z_k + d_k^z$, $s_{k+1} \gets s_k + d_k^s$
\EndFor \\

\Return $(x_k, y_k, z_k, s_k)$

\end{algorithmic}
\end{algorithm}

\section{Numerical Experiments}
\label{sec:num-exp}
In this section, we provide a comprehensive evaluation of the proposed method.

\subsection{System setup}

\subsubsection{Cost function design}
We consider the quadratic cost function on $\mathrm{SO}(3) \times \mathbb{R}^3$ to indicate the difference between the desired and the current configurations. For orientation and the discrete angular velocity, we consider the square of the weighted chordal distance~\cite{lee2010geometric}:
\begin{equation}
    L_{\mathrm{SO}(3)}(R, R_d) = \frac{1}{2}\operatorname{tr}\left( (RR^{\transpose}_d - I)W_R(RR^{\transpose}_d - I)^{\transpose} \right),
\end{equation}
where $W_R$ is a positive-definite weighting matrix.
The derivative of the cost function can be obtained by considering the retraction curve:
\begin{equation}
    L_{\mathrm{SO}(3)}(R\exp{(t\xi)}, R_d), \xi \in \mathfrak{so}(3). 
\end{equation}
For distance, we consider the quadratic cost function in $\mathbb{R}^3$:
\begin{equation}
    L_{\mathbb{R}^3}(p, p_d) = \frac{1}{2}(p - p_d)^{\transpose}W_p(p - p_d).
\end{equation}
where $W_p$ is a positive-definite weighting matrix. 
\subsubsection{Software setup}
The main loop of \Cref{alg:line-search-RIPM} is implemented in MATLAB with the internal linear system solvers for the Newton steps. We use CasADi\cite{andersson2019casadi} to evaluate the second-order retraction curve in \Cref{table:cons_2nd_retraction} to obtain the first- and second-order derivatives. To improve the performance, we use code generation to obtain the C++ file and compile it in MATLAB as MEX files for execution. We consider CasADi as our interface for implementing direct trajectory optimization for the baselines using IPOPT and YALMIP \cite{Lofberg2004} when using SNOPT \cite{gill2005snopt}. Due to the difference between the evaluation of the derivatives, we compare the time spent in the solver and the number of iterations as the metric. 

\begin{figure}[t]
    \centering
    \includegraphics[width=1\linewidth]{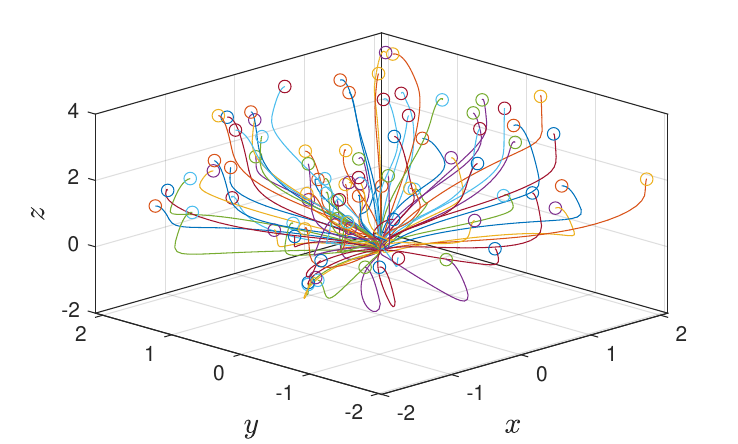}
    \caption{Testing cases for drone docking from randomly generated initial poses in $\mathrm{SO}(3)\times \mathbb{R}^3$.}
    \label{fig:srb-cases}
\end{figure}

\begin{figure*}[t]
    \centering
    \includegraphics[width=2\columnwidth]{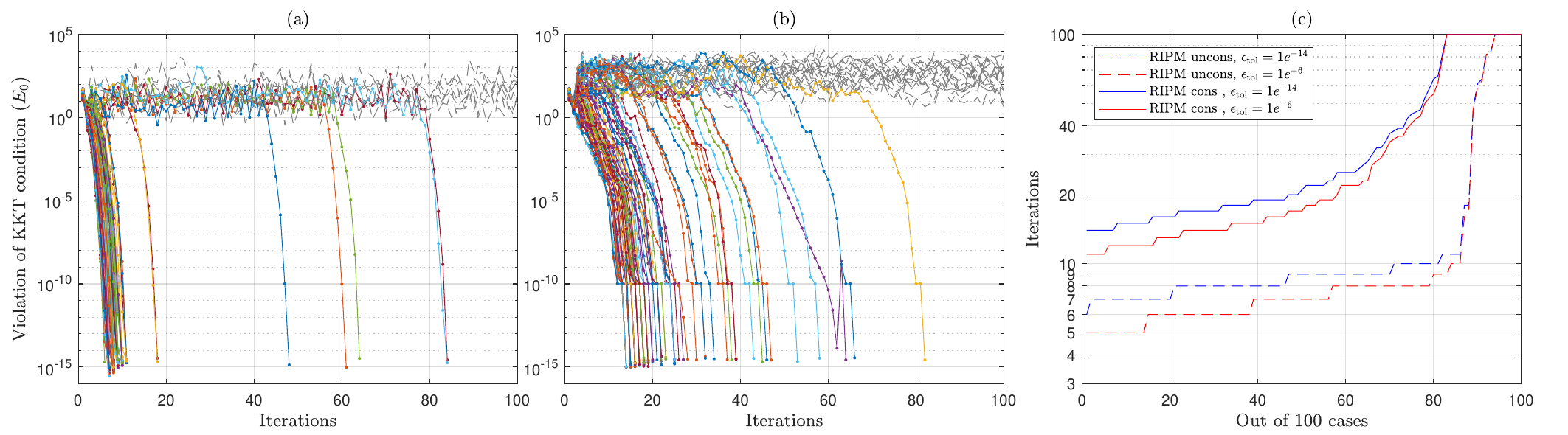}
    \caption{Convergence of RIPM on drone on $\mathrm{SO}(3)\times \mathbb{R}^3$ with full dynamics with $100$ different initial conditions. The tolerance for convergence is set to $E_0 = 1e^{-14}$. The convergence rate of unconstrained cases in (a) is much faster than that of constrained cases in (b). The dashed lines indicate the cases that do not converge, while the others are converged cases. Accelerated convergence or super-linear convergence is observed to indicate the success of the second-order method. In (c), we compared the iterations to converge. We can see that it takes fewer iterations for both cases to stop with the default tolerance of IPOPT $\epsilon_{\mathrm{tol}} = 1e^{-6}$. The convergence rate and speed outperform the quaternion-based method IPOPT computed in the ambient space.
    }
    \label{fig:ripm-srb-converge}
\end{figure*}

\subsubsection{Hardware setup} We use a laptop with an Intel i7-11850H CPU to run all the experiments.

\subsection{Single Rigid Body}
In the first experiment, we demonstrate the proposed method on a single rigid body. We consider a quadrotor modeled as a full rigid body with input limits. The input is assumed to be a 4-DOF vector composed of the total linear force along the $z-$axis in the body frame and a 3-DOF torque. As a result, the dynamics part can be expressed as:
\begin{equation}
\label{eq:drone_dynamics}
\begin{aligned}
    &F_{k+1}I^b - I^bF_{k+1}^{\transpose}=I^bF_{k} - F_k^{\transpose}I^b + \tau^{\wedge} \Delta t^2 , \\ 
    &mv_{k+1} = mv_{k} + mg\Delta t + R_{k+1}e_zu_z\Delta t. \\ 
\end{aligned}
\end{equation}
We assume that the torque admit a symmetric box limits $\tau \in [-\tau_{\lim}, \tau_{\lim}]^3 $ and the linear force is within a non-negative interval $u_z \in [0, u_{z, \lim}]$.

As the minimal dimension of a singularity-free representation of $\mathrm{SO}(3)$ is 4, we do not consider the 3-DOF parameterization, such as Euler angle, Rodriguez formula, or the modified quaternion representation \cite{brudigam2021integrator}. We compare the proposed formulation with a sigularity-free quaternion-based variational integrator.
As the quaternion parameterizes the rotation with a four-dimensional vector
$q := \begin{bmatrix}
        q_v, q_x, q_y, q_z
    \end{bmatrix} \in \mathbb{R}^4$,
it does not have singularities. However, the constraint $\|q\|^2 = 1$ is required to ensure $q \in \mathbb{S}^3$, and $q_v \ge 0$ is necessary to ensure the state lies in one branch of the manifold. Consider the integrator \cite{brudigam2021integrator}\footnote{The original implementation of \cite{brudigam2021integrator} considers $q_v = \sqrt{1-q_x^2-q_y^2-q_z^2}$ to avoid manifold constraints and reduce dimensions, which, however, introduces a singularity that results in \texttt{NaN} error in numerical optimizations.} 
with the singularity-free representation of quaternions, we directly apply the CasADi and its default IPOPT solver to conduct trajectory optimization with the gradient derived in the ambient space. 

We consider the task of docking the drone from randomly sampled initial poses to the origin with an identity pose and zero twists as shown in \Cref{fig:srb-cases}. The location of the drone is uniformly sampled from a box. The orientation is also randomly sampled such that $\log{(R_0)}$ is uniform in the interval $[0, \pi]$ and all directions. We apply an initialization strategy that uniformly interpolates the geodesic curves between the initial and the goal states. 
The convergence result of the RIPM is presented in \Cref{fig:ripm-srb-converge}. As we can see, 82 out of 100 cases converge while 93 out of 100 of the unconstrained cases converge to the tolerance $\epsilon_{\mathrm{tol}} = 1e^{-14}$. The ability to converge super-linearly to such high precision demonstrates the power of the proposed second-order method. For the constrained cases, more than $50\%$ cases converge in 50 iterations if the tolerance is set to the default value of IPOPT $1e^{-6}$. 

\begin{table*}[t]
\centering
\caption{Comparison of convergence rate of single rigid body case. We summarize the convergent cases for the proposed method and baselines. Otherwise noted, all the cases are without input inequality constraints. Without certified optimal solutions (SDP), the baseline can not converge for either IPOPT with exact Hessian or SNOPT with approximated Hessian. With certified initialization, more than half of the cases using IPOPT can converge, while the performance degrades soon after noises are present. Similar performance is observed for SNOPT, which is even more sensitive to noise. The time in each iteration for the IPM-based method is also listed. The proposed method also takes only about $30\%$ of the time as in IPOPT for the unconstrained cases in each iteration. The planning horizon is $40$ steps.}
\label{table:quat-converge}
\begin{tabular}{cccccc}
\hline
\multirow{2}{*}{Methods} & \multirow{2}{*}{Initialization Strategy} & Convergence in $1000$ Iters. & \multicolumn{2}{c}{Iters. to Convergence} & Avg Time per Iteration ($\sec$)\\ 
& &(in $100$ Iters. for proposed) & Mdn & Avg & w/o function evaluation\\
\hline
\multirow{2}{*}{Proposed} & \multirow{2}{*}{Straight line} & $82/100$ (constrained)& $19$ & $24$ & $0.0078$\\
 &  & $93/100$ & $9$ & $12$ & $0.0046$\\
\hline 
\multirow{4}{*}{IPOPT} & Straight line & $0$ & $-$ & $-$ & $0.0136$\\ 
& SDP & $68/100$ & $5$ & $73$ & $0.0133$ \\ 
& SDP($\Sigma_1$) & $45/100$ & $106$ & $211$ & $0.0113$\\ 
& SDP($\Sigma_2$) & $31/100$ & $317$ & $382$ & $0.0101$ \\ 
\hline 
\multirow{4}{*}{SNOPT} & Straight line & $0$ & $-$ & $-$ & $-$\\ 
& SDP & $56/100$ & $170$ & $177$ & $-$ \\ 
& SDP($\Sigma_1$) & $1/100$  & $975$ & $975$ & $-$\\ 
& SDP($\Sigma_2$) &  $0$  & $-$ & $-$ & $-$\\ 
\hline 
\end{tabular}
\end{table*}

We also present the result of the quaternion-based method in \Cref{table:quat-converge}. We consider the baseline solver as the IPOPT based on the Interior Point Method and the SNOPT \cite{gill2005snopt} based on Sequential Quadratic Programming (SQP).  Due to the polynomial nature of the dynamics and the manifold constraints, most cases do not converge in 1000 iterations with the baselines. As the gradient-based nonlinear programming is sensitive to initializations, we also provide high-quality initialization from the globally optimal motion planning methods \cite{teng2024convex} based on moment relaxation (dual of sum-of-square optimization)\cite{lasserre2001global, lasserre2015introduction}. We apply the open-source code in \cite{teng2024convex} to generate a certified optimal trajectory and perturb the solutions with different levels of Gaussian noise as initialization. We perturbed the position and velocity of the SDP-based solution by Gaussian noise with covariance $0.1I$. We perturb the orientation and the discrete angular velocity on a quaternion with Gaussian noise with covariances $\Sigma_1 = 0.01I$ and $\Sigma_2 = 0.05I$, project them to $\mathbb{S}^3$ by rounding the norm to $1$. As shown in \Cref{table:quat-converge}, the convergence rate of IPOPT and SNOPT soon degrades when the perturbation increases. 

\begin{figure*}
    \centering
    \includegraphics[width=1\linewidth]{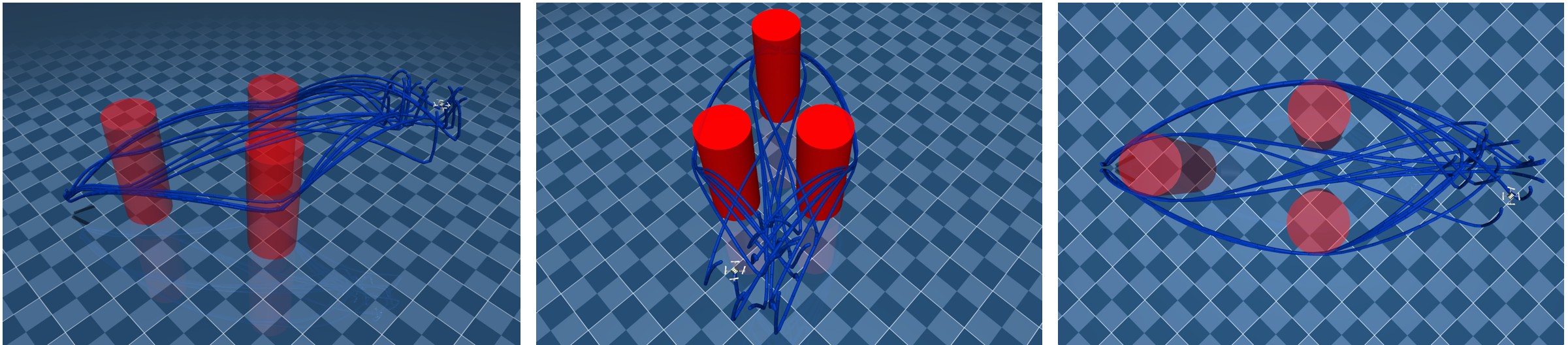}
    \caption{Trajectories optimization for drone traversing complex environments considering general nonlinear constraints.}
    \label{fig:drone-cluttered}
\end{figure*}

Finally, we show that the proposed method is capable of handling general nonlinear constraints by designing trajectories for the drone in cluttered environments as shown in \Cref{fig:drone-cluttered}. We consider nonconvex norm constraints $(x - x_c)^2 + (y - y_c)^2 \ge r^2$ for the drones to avoid the cylindrical regions. As the proposed RIPM considers general nonlinear constraints, future work can also incorporate more sophisticated representations of collision-free regions, such as \cite{amice2023finding, wu2025towards}. 

\subsection{Multi-Rigid Body}

\begin{figure}[th]
    \centering
    \includegraphics[width=1\linewidth]{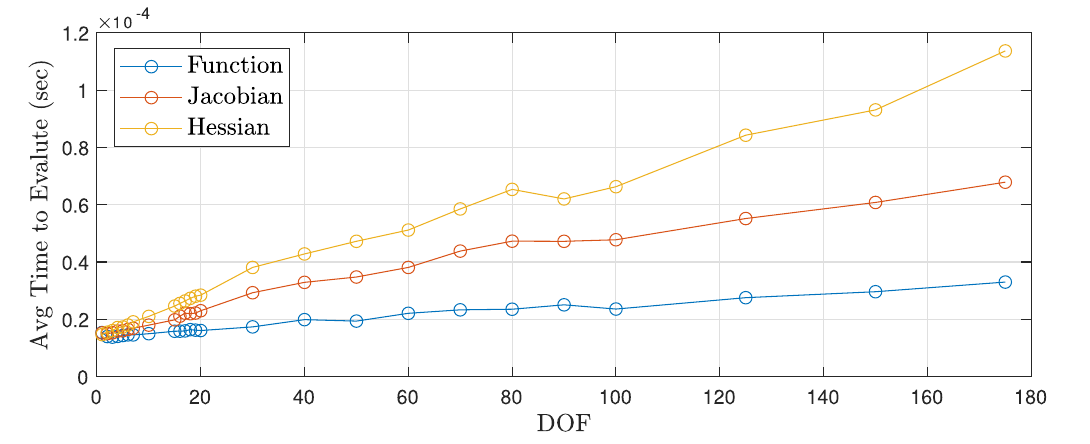}
    \caption{Average time to evaluate the zeroth-, first-, and second-order derivative of rigid body dynamics w.r.t. the DOF of a kinematic chain with one branch. The proposed formulation scales linearly w.r.t. the DOF. }
    \label{fig:complexity-dof}
\end{figure}

\begin{figure*}[t]
    \centering
    \includegraphics[width=0.9\linewidth]{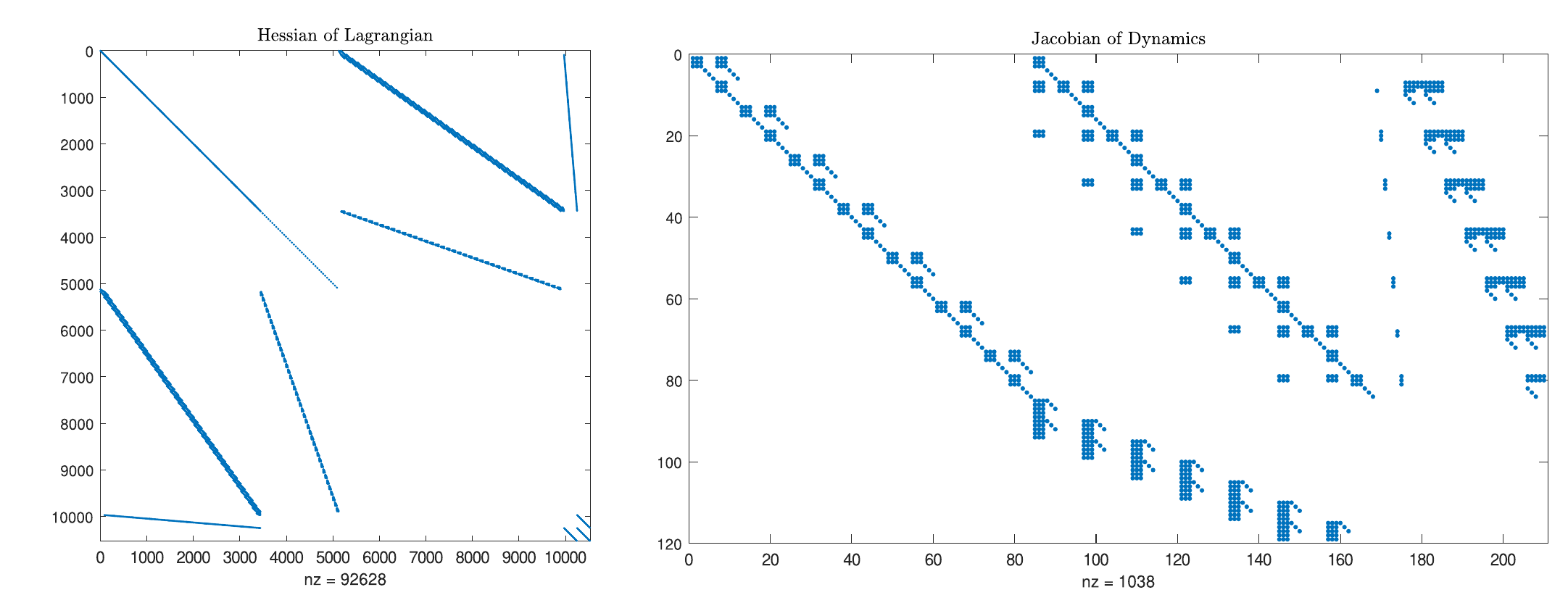}
    \caption{Illustration of the Hessian of the Lagrangian and Jacobians of the rigid body dynamics of the manipulator example. The block diagonal structure shows that the KKT system scales linearly w.r.t. the planning horizon. The planning horizon is 40 steps for the manipulator task using full dynamics. }
    \label{fig:sparsity}
\end{figure*}

\textbf{Complexity analysis}: We verify the complexity of the proposed differentiation via a kinematic chain on $\mathrm{SO}(3)\times \mathbb{R}^3$ with a single branch and different depths. We compare the time to evaluate the zeroth- to the second-order differentiation of the dynamics. As is shown in \Cref{fig:complexity-dof}, the time scales linearly w.r.t. the depth of the kinematic chain. The sparsity pattern of a kinematic chain and the Hessian of the Lagrangian of the following example is shown in \Cref{fig:sparsity}. The block diagonal structure indicates that the KKT system scales linearly w.r.t. the planning horizon.

\begin{table*}[t]
\centering
\caption{Comparison of proposed formulation using RIPM on $\mathcal{M}_{\mathrm{RB}}$ with IPOPT in generalized coordinates. Without infeasibility restoration, the proposed method has a $25\%$ convergence rate. With the infeasibility restoration, IPOPT greatly outperforms the proposed method at the cost of more iterations to solve feasibility sub-problems. In terms of iterations to converge, the proposed method greatly outperforms IPOPT, while the time consumed in each iteration is three times higher, due to the increasing number of variables. }
\label{table:multi-converge}
\begin{tabular}{cccccc}
\hline
\multirow{2}{*}{Methods} & Convergence in $500$ Iterations & \multicolumn{3}{c}{Iters. to Convergence} & Time per Iteration ($\sec$)\\ 
& (w/o infeasibility restoration)  & Mdn & Avg &Max & w/o function evaluation\\
\hline
{RIPM $(\mathcal{M}_{\mathrm{RB}})$ } & $21/100$ & $21$ & $20$ &$42$ & $0.1159$\\  \hline

\multirow{2}{*}{IPOPT $(q\in \mathbb{R}^{n})$} & $64/100$ & $147$ & $201$ & $498$ & $0.0319$\\
 & $(16/100)$ & $57$ & $76$ & $197$ & $0.0329$\\ 
\hline 
\end{tabular}
\end{table*}


\textbf{Motion planning of manipulator}: In the second example, we apply the proposed method to motion planning of a 7-DOF KUKA iiwa manipulator. We model each link of the manipulator and enforce the revolute joint constraints via considering one pivot constraint and a 2-DOF axis constraint as shown in \Cref{table:cons_2nd_retraction}. We consider the baseline as the direct methods with full dynamics in the generalized coordinates:
\begin{equation}
\begin{aligned}
    &q_{k+1} = q_k + \dot{q}_k\Delta t, \\
    &M(q_{k})\frac{(\dot{q}_{k+1} - \dot{q}_k)}{\Delta t} + C(q_k, \dot{q}_k)\dot{q}_k +G(q_k) = \tau_k.
\end{aligned}
\end{equation}
We consider moving the manipulator from the initial pose to the target pose while avoiding obstacles, a cylinder with fixed location and radius. For simplification, we only penalize the distance between the center of mass of each link and the obstacle. Thus, the proposed method will have constraints for the $i-th$ link as: 
\begin{equation}
    (x_i - x_{\mathrm{obs}})^2 + (y_i - y_{\mathrm{obs}})^2 \ge r_{\mathrm{obs}}^2, i = 1, 2, \cdots,7.  
\end{equation}
For generalized coordinates formulation, forward kinematics is required to map the joint angle $q \in \mathbb{R}^n$ to the task space. 

We initialize the trajectory via linear interpolation of joint space in both cases. As the manipulator is fully actuated, we can warm start the trajectory with feasible primal solutions by computing the constrained force and the input torque through inverse dynamics. In all cases, we consider a fixed target pose with a random initial pose and a fixed obstacle and target pose. The convergence rates are shown in \Cref{table:multi-converge}. As IPOPT admits advanced features like infeasibility restoration that is not applied in this method, we list the results of IPOPT with and w/o this feature for fair comparison. We find that our method outperforms IPOPT w/o infeasibility restoration in terms of convergence rate and number of iterations to converge. With infeasibility restoration, the performance of IPOPT improves at the cost of more iterations to converge. For both cases, we consider $\epsilon_{\mathrm{tol}} = 1e^{-6}$ as the termination condition. 






\section{Limitations and Discussions}
\label{sec:lim}
Here, we analyze the limitations of the proposed algorithm. 

\textbf{Increasing number of variables}: The proposed method considers all the degrees of freedom of the rigid body, thus making the KKT system multiple times larger than the version in the generalized coordinates. Thus, the trade-off between the number of iterations and computational time in each iteration should be carefully investigated. First-order Riemannian optimization \cite{liu2020simple} can be considered in future work to bypass the need to solve the expensive Newton steps. The discrete null space mechanics \cite{betsch2005discrete, betsch2006discrete, leyendecker2010discrete} can also be applied to eliminate the constrained force, while the null space of the Jacobians of constraints needs to be carefully computed. 

\textbf{Guarantees on global optimality}: The proposed method is based on local gradients. As a result, global optimality is not guaranteed. Conducting optimization landscape analysis is also challenging due to the nonconvex nature of motion planning problems. However, we note that $\mathrm{SO}(3)\times \mathbb{R}^3$ is constrained by polynomial quality, thus can be seamlessly integrated with the globally optimal motion planning framework using moment relaxations \cite{teng2024convex, kang2024fast, yang2022inexact}. 

\textbf{Holonomic constraints}: The proposed method only considers the holonomic constraints, while it does not incorporate more complicated structures, such as complementarity constraints \cite{posa2014direct}, nonholonomic constraints \cite{bloch2003nonholonomic}, or the hybrid zero dynamics \cite{westervelt2003hybrid}. As the RIPM is a general framework for constrained optimization, we plan to include these constraints by taking these special structures into consideration \cite{kobilarov2010geometric, howell2022calipso, hereid2017frost}.  

\textbf{Code optimization}: The proposed RIPM is implemented in MATLAB and does not fully replicate all the features of IPOPT \cite{wachter2006implementation}. As we do not include the advanced features, such as inertial correction, second-order correction, and infeasibility restoration, the proposed method is not as robust as IPOPT. It is crucial to include these features to improve the performance of the proposed solver further. 

\section{Conclusions}
\label{sec:conclusion}

In this work, we, for the first time, applied the constrained Riemannian optimization to the paradigm of direct trajectory optimization, which is fast and preserves the topological structure of rotation groups. We first model the multi-body dynamics on matrix Lie groups using the Lie group variational integrator. Then we derive the closed-form Riemannian derivative and apply the line-search Riemannian Interior Point Method to conduct trajectory optimization. We demonstrate that the complexity in the derivative evaluation and Newton steps is linear w.r.t. both the system DOF and the planning horizon. We also show that the proposed method outperforms the conventional methods by an order of magnitude in challenging robotics tasks. 



\section*{Acknowledgments}
W. Clark, R. Vasudevan, and M. Ghaffari were supported by AFOSR MURI FA9550-23-1-0400. The authors thank the area chairs and anonymous reviewers for the constructive feedback and in-depth review.

\begin{appendices}
\section{Derivation of second-order retraction of kinematic chain}
\label{appx:kin-derivation}

Consider the vectorized kinematic chain \eqref{eq:vec-kin}:
\begin{equation*}
    \log{(\bar{Y}^{-1}X_1X_2\cdots X_{n-1}X_n)} = \log{(\bar{Y}^{-1})}. 
\end{equation*}
Denote $X_{i, j} = X_i X_{i+1} \cdots X_j$ when $i \le j$ and $X_{i+1, i} = I$, we have the identity using the property of adjoint action:
\begin{equation}
    \begin{aligned}
    \exp{(t\xi)\bar{X}_{i,j}} 
    &= \bar{X}_{i,j} (\bar{X}_{i,j}^{-1} \exp{(t\xi)} \bar{X}_{i,j}) \\ 
    &= \bar{X}_{i,j} \exp{(t \operatorname{Ad}_{\bar{X}_{i, j}^{-1} } \xi)}. 
\end{aligned}
\end{equation}
The perturbed kinematic chain becomes:
\begin{equation}
\begin{aligned}
    &Y^{-1}\bar{X}_{1}\exp{(t\xi_1)}\cdots\bar{X}_{n-1}\exp{(t\xi_{n-1})} \bar{X}_{n}\exp{(t\xi_n)}\\
    =&Y^{-1}\bar{X}_{1}\exp{(t\xi_1)} \cdots \\
    &\bar{X}_{n-1} \bar{X}_n \exp{(t \operatorname{Ad}_{\bar{X}_{n, n}^{-1} } \xi_{n-1})}\exp{(t\xi_n)}\\
     =&Y^{-1}\bar{X}_{1}\exp{(t\xi_1)} \cdots  \bar{X}_{n-2} \\
     & \bar{X}_{n-1} \bar{X}_n  (\bar{X}_{n}^{-1} \bar{X}_{n-1}^{-1} \exp{(t\xi_{n-2})} \bar{X}_{n-1} \bar{X}_n) \\
     & \exp{(t \operatorname{Ad}_{\bar{X}_{n, n}^{-1} } \xi_{n-1})}\exp{(t\xi_n)}\\
     =&Y^{-1}\bar{X}_{1}\exp{(t\xi_1)} \cdots  \bar{X}_{n-2}\bar{X}_{n-1} \bar{X}_n  \exp{(t \operatorname{Ad}_{\bar{X}_{n-1, n}^{-1} })}\\
     & \exp{(t \operatorname{Ad}_{\bar{X}_{n, n}^{-1} } \xi_{n-1})}\exp{(t\xi_n)}\\
    =&Y^{-1}\bar{X}_{1,n}\exp{(t \operatorname{Ad}_{\bar{X}_{2, n}^{-1} } \xi_{1})}\exp{(t \operatorname{Ad}_{\bar{X}_{3, n}^{-1} } \xi_{2})} \cdots \exp{(t\xi_n)} \\
    =&\exp{(t \operatorname{Ad}_{\bar{X}_{2, n}^{-1} } \xi_{1})}\exp{(t \operatorname{Ad}_{\bar{X}_{3, n}^{-1} } \xi_{2})} \cdots \exp{(t\xi_n)}.
\end{aligned}
\end{equation}
Then we proceed to apply the BCH formula to obtain the second-order retraction:
\begin{equation}
\begin{aligned}
     &\log{\left( \prod_{i=1}^n \exp{\left( t\operatorname{Ad}_{\bar{X}^{-1}_{i+1, n}} \xi_i \right)} \right)} \\
    =& t\operatorname{Ad}_{\bar{X}^{-1}_{2, n}} \xi_1 + \log{\left( \prod_{i=2}^n \exp{\left( t\operatorname{Ad}_{\bar{X}^{-1}_{i+1, n}} \xi_i \right)} \right)} \\
    +&  \frac{1}{2} \left[t\operatorname{Ad}_{\bar{X}^{-1}_{2, n}} \xi_1, \log{\left( \prod_{i=2}^n \exp{\left( t\operatorname{Ad}_{\bar{X}^{-1}_{i+1, n}} \xi_i \right)} \right)}\right] + \mathcal{O}(t^3)
\end{aligned}
\end{equation}
By repeated application of BCH formula to $\log{\left( \prod_{i=k}^n \exp{\left( t\operatorname{Ad}_{\bar{X}^{-1}_{i+1, n}} \xi_i \right)} \right)}, k\ge 2$, we have the second-order retraction as: 
\begin{equation}
     \begin{aligned}
         & \quad \quad \log \left( \prod_{i=1}^n \exp{\left( t\operatorname{Ad}_{\bar{X}^{-1}_{i+1, n}} \xi_i \right)} \right) = \\
          &t\sum_{i=1}^n \operatorname{Ad}_{X_{i+1, n}^{-1}} \xi_i + \sum_{i < j} \frac{t^2}{2}[\operatorname{Ad}_{X_{i+1, n}^{-1}} \xi_i, \operatorname{Ad}_{X_{j+1, n}^{-1}} \xi_j] + \mathcal{O}(t^3)
     \end{aligned}
\end{equation}
We note that the derivation does not involve the Jacobian of the matrix logarithmic map due to the shift of the operating point by $\bar{Y}$. 

\section{Implementation details of RIPM}
\label{appx:ripm-table}
We consider the following parameters shown in Table~\ref{table:param-RIPM} to implement the RIPM in \Cref{alg:line-search-RIPM}.
\begin{table}
\centering
\caption{Part of hyper-parameters in \Cref{alg:line-search-RIPM}.}
\label{table:param-RIPM}
\begin{tabular}{ccc}
\hline
 Parameter & Notation & Value \\ \hline
Maximal iteration & ${N}_{\max}$ &  $200$ \\ \hline
Maximal iteration in line search & $J_{\max}$ &  $30$ \\ \hline
Tolerance to KKT of \Cref{prob:cons-nlp} & $\epsilon_{\mathrm{tol}}$ &  $1e^{-11}$ \\ \hline
Linear decaying rate of $\mu$ & $\kappa_{\mu}$ &  $0.99$ \\ \hline
Superlinear decaying rate of $\mu$ & $\theta_{\mu}$ &  $1.99$ \\ \hline
Minimal fraction to boundary & $\tau_{\min}$ &  $0.995$ \\ \hline
Barrier problem cost progress & $\gamma_{\theta}$ &  $1e^{-6}$ \\ \hline
Minimal infeasibility & $\theta_{\min}$ &  $1e^{-4}$ \\ \hline
Progress of barrier cost & $\eta_{\varphi}$ &  $1e^{-4}$ \\ \hline
Progress of feasibility & $\gamma_\theta$ &  $1e^{-4}$ \\ \hline
Decaying rate for line search & $\beta$ & $0.5$ \\ \hline  
\end{tabular}
\end{table}

\end{appendices}
{
\small 
\balance
\bibliographystyle{unsrtnat}
\bibliography{bib/strings-full,bib/ieee-full,bib/references}
}

\end{document}